%% file: neurips_2020.tex
\documentclass{article}

\usepackage[pagebackref=true,breaklinks=true,letterpaper=true,colorlinks,bookmarks=false]{hyperref}

\usepackage[nonatbib,preprint]{neurips_2020}

\usepackage[utf8]{inputenc} 
\usepackage[T1]{fontenc}    
\usepackage{hyperref}       
\usepackage{url}            
\usepackage{nicefrac}       
\usepackage{microtype}      
\usepackage{cite}
\usepackage{graphicx,float}
\usepackage[symbol]{footmisc}
\usepackage[flushleft]{threeparttable}

\usepackage{subcaption}
\usepackage{amsmath,amsfonts,amsthm,amssymb}
\usepackage{booktabs, multirow, array}
\DeclareMathOperator*{\argmax}{arg\,max}

\DeclareMathOperator*{\argmin}{arg\,min}
\newcommand{\PreserveBackslash}[1]{\let\temp=\\#1\let\\=\temp}
\newcolumntype{C}[1]{>{\PreserveBackslash\centering}p{#1}}
\newcolumntype{L}[1]{>{\PreserveBackslash}p{#1}}
\newcolumntype{R}[1]{>{\PreserveBackslash\flushright}p{#1}}

\usepackage[normalem]{ulem} 
\usepackage{xcolor} 

\newcommand{\del}[1]{}  
\newcommand{\note}[1]{}
\newcommand{\rev}[1]{\textcolor{black}{{#1}}}

\title{Tackling Multiple Tasks with One Single Learning Framework}

\author{%
  Michael Yang\\
  OPPO\\
  \texttt{xuewen.yang@protonmail.com} \\
}

\begin{document}

\maketitle

\begin{abstract}
Deep Multi-Task Learning (DMTL) has been widely studied in the machine learning community and applied to a broad range of real-world applications. Searching for the optimal knowledge sharing in DMTL is more challenging for sequential learning problems, as the task relationship will change in the temporal dimension. In this paper, we propose a flexible and efficient framework called Hierarchical-Temporal Activation Network (HTAN) to simultaneously explore the optimal sharing of the neural network hierarchy (hierarchical axis) and the time-variant task relationship (temporal axis). HTAN learns a set of time-variant activation functions to encode the task relation. A functional regularization implemented by a modulated SPDNet and adversarial learning is further proposed to enhance the DMTL performance. Comprehensive experiments on several challenging applications demonstrate that our HTAN-SPD framework outperforms SOTA methods significantly in sequential DMTL.
\end{abstract}

\input{section1}
\input{section2}
\input{section3}

\input{section4}
\input{section5}

\input{section6}

\newpage
\bibliography{neurips_2020} 
\bibliographystyle{IEEEtran}

\input{section_appendix}

\end{document}

%% file: section1.tex
\section{Introduction}
Multi-Task Learning (MTL) is a widely studied field in machine learning community, where models for a set of related tasks are trained simultaneously to improve the performance \cite{Caruana93ICML,Andrew2017ICML,Zhao2018AAAI,Zhang2018NIPS,Yousefi2019NIPS}. Deep Multi-Task Learning (DMTL) further incorporates the flexibility of deep learning models into MTL paradigm and has achieved remarkable performance in many applications, ranging from Computer Vision \cite{kendall2018CVPR,Sener2018NIPS,Liu2019CVPR,Lee2019CVPR} to Natural Language Processing \cite{Feng2018EMNLP,Sanh2019AAAI,Nishino2019EMNLP}. The central problem in DMTL is to model the task-specific and shareable knowledge across different tasks, which is encoded in the architecture and parameters of the neural network. For non-sequential data such as images, existing DMTL methods can be categorized two strategies: hard-sharing and soft-sharing. The hard-sharing methods require manually defined sharing architecture for all the tasks. Soft-sharing methods define individual network for each task and learn the task relation by either regularization or sub-module connection. While hard-sharing models are unable to automatically discover the optimal shared architecture, soft-sharing models are capable of learning the knowledge sharing from data. However, the model complexity of soft-sharing models explodes with respect to the number of tasks.

DMTL for sequential learning problems arises naturally in many real-world applications \cite{Pengfei2017ACL,Feng2018EMNLP,Sanh2019AAAI,Nishino2019EMNLP,Divam2019AAAI,Wu2019EMNLP,Dankers2019EMNLP,Liu2019ACL,Chen2018AAAI,Zare2018ACL}. DMTL for sequential data is much more challenging than non-sequential ones, as the knowledge sharing across tasks can be also time-variant. A example is given as follows. We visualize the absolute values of covariances between the POS tags and named entities in the OntoNote 5.0 dataset\footnote{https://catalog.ldc.upenn.edu/LDC2013T19} with respect to the time slots. \rev{In each time slot $n$, we analyze the occurrence of POS tag $Y_{n,1}$ and named entity $Y_{n,2}$. The covariance is computed as $\text{COV}(Y_{n,1},Y_{n,2})=\mathbb{E}(Y_{n,1}Y_{n,2}) - \mathbb{E}(Y_{n,1})\mathbb{E}(Y_{n,2})$, where $\mathbb{E}$ denotes the average in the whole dataset.} The amplitude variation indicates that the relationship between the two tasks (i.e. POS tagging and named entity recognition) changes over time. \rev{Therefore, the shared architecture of the deep learning model should also evolve through time.} In spite of the great achievement, existing DMTL methods for sequential data generally ignore this time-invariant nature of the task dependency, which will leads to inaccuracy on the task performance. 

\vspace{-0.3cm}
\begin{figure}[H]
	\centering
	\includegraphics[clip, width=0.70\linewidth]{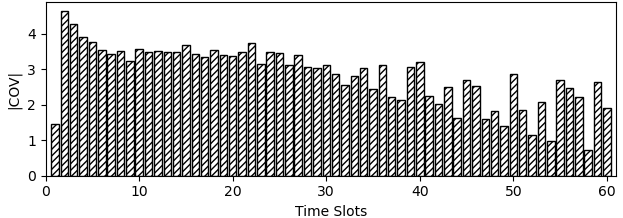}
	
	\caption{Absolute value of the covariance between POS tags and named entities in different time}
	\label{Fig:factor}
\end{figure}
\vspace{-0.5cm}

In this paper, we propose a sequential DMTL framework called Hierarchical-Temporal Activation Network (HTAN) to explore the time-variant task relationship for sequential data. HTAN learns task-specific time-variant activation functions and embeds shared knowledge by the similarity between the functions. Based on the similarity metrics of the activation functions, we further incorporate a modulated SPDNet and propose a functional regularization method to enhance the performance of HTAN. Compared with literature works, our HTAN-SPD framework is able to flexibly capture the time variant task relationship and learn better shared architecture.

%% file: section2.tex
\section{Related Works}
\subsection{Sequential Deep Multi-Task Learning}

The most straight-forward way for sequential DMTLs is the hard-sharing strategy \cite{Liu2019ACL,Nishino2019EMNLP,Ning2017AAAI,He2019ACL,Nishida2019ACL,Wei2019ACL,Dou2019EMNLP}, which is unable to extract the optimal shared architecture from data. 
In sequential DMTL for NLP applications, there are also some soft-sharing studies \cite{Feng2018EMNLP,Pengfei2017ACL,Zare2018ACL,Divam2019AAAI,Wu2019EMNLP,Yang2019AAAI,Shafiq2018EMNLP,pentyala2019ACL}, where individual neural network is defined for each task. One typical soft-sharing methodology is training totally $T + 1$ sequential modules to learn task-specific and shareable features separately \cite{Pengfei2017ACL,Zare2018ACL,Divam2019AAAI,Wu2019EMNLP,pentyala2019ACL,lin2018ACL,Dankers2019EMNLP}, where $T$ is the number of tasks. The other representative method is using Meta Learning \cite{Chen2018AAAI}, where a shared meta network is incorporated to modulate the parameters of task-specific networks. Although these methods are more flexible than hard-sharing methods, their model complexity explodes with respect to the number of tasks, which makes them not scalable for applications with a large amount of related tasks. 

To the best of our knowledge, the majority of existing works is unable to capture the time-variant task relations, As the task relation decides how much knowledge should be shared across tasks at different time, using time-invariant shared architecture will weaken the model performance. In some time slots, the model may share too much knowledge across unrelated tasks. But in other time slots, the model may not fully explore the shareable knowledge between highly dependent tasks. The most closest methods of our works are GIRNet \cite{Divam2019AAAI}, Meta-LSTM \cite{Chen2018AAAI} and MTL-Routing \cite{Zare2018ACL}. Although they can probably learn time-variant task relation, our proposed method is more flexible and has smaller model complexity under certain condition.

\subsection{Learning Activation Functions}
Our proposed HTAN shares some similar intuitions with activation function learning. In order to extend the capability of deep learning models, there are multiple studies on learning more flexible activation functions \cite{Agostinelli2015ICLR,Hou2017AISTATS} for neural networks. The authors \rev{propose a TAAN network to} incorporate the activation function learning into feed-forward networks for flexible and low-cost DMTL. TAAN further introduces a quadratic term of $\alpha^{t}_{l}$ as the functional regularization term. Compared with our proposed method for sequential DMTL, there are several disadvantages of TAAN and its functional regularization. First, TAAN is proposed for non-sequential data. Its activation functions can not evolve with respect to time. Secondly, \rev{there is a positive definite matrix in the regularization, which is computed by assuming the input follows standard Gaussian distribution. In sequential data, this assumption is generally violated and will lead to inaccuracy.}

\vspace{-0.3cm}
\begin{figure}[H]
	\centering
	\begin{subfigure}[t]{.30\linewidth}
	\includegraphics[clip, width=0.90\linewidth]{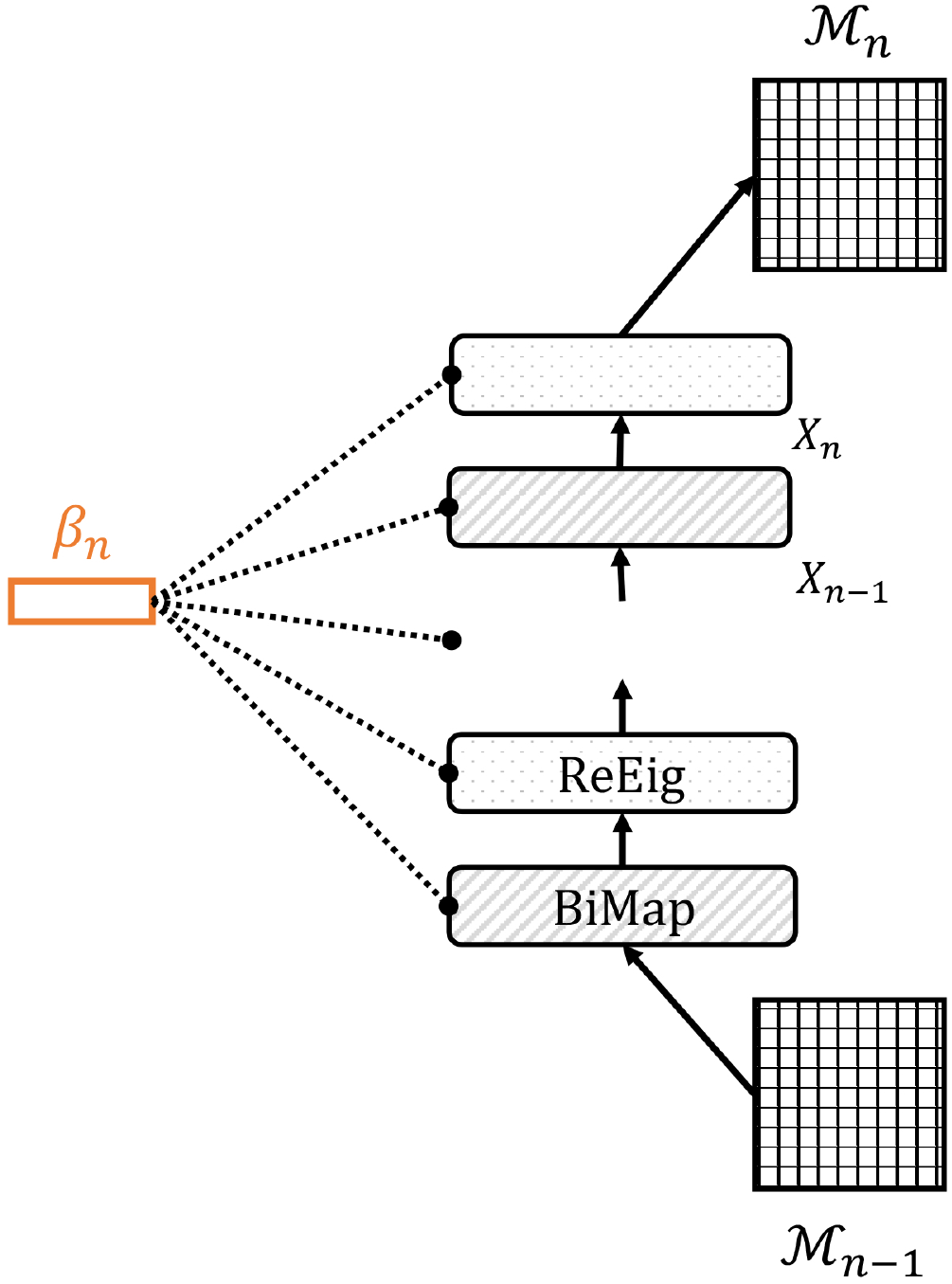}
	\caption{}
	\end{subfigure}\hfill
	\begin{subfigure}[t]{.60\linewidth}
	\includegraphics[clip, width=0.85\linewidth]{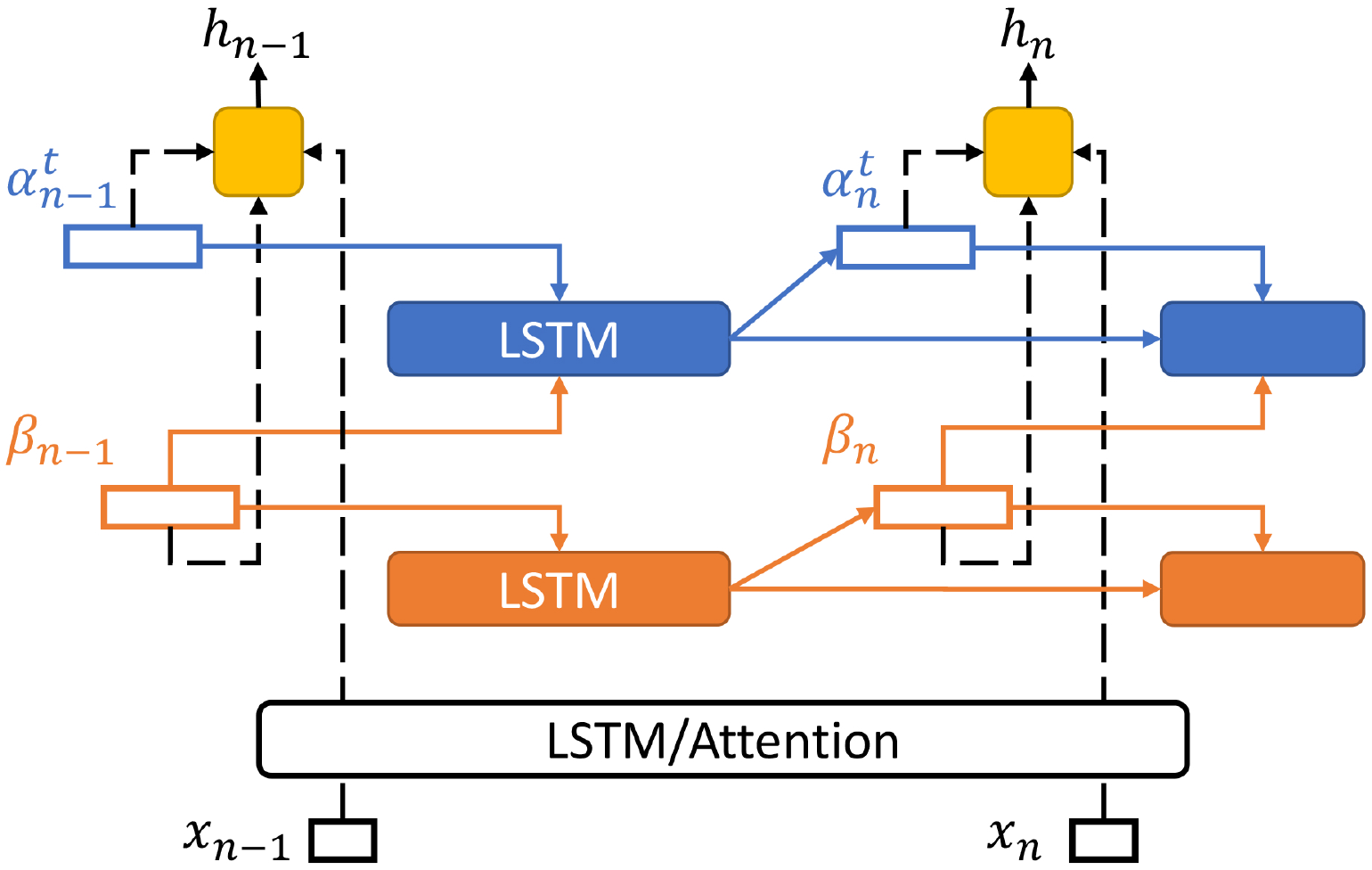}
	\caption{}
	\end{subfigure}
	\caption{(a) Architecture of $\beta_n$-Modulated SPDNet; (b) Structure of task adaptive block in HTAN}
	\label{Fig:HTAN}
\end{figure}
\vspace{-0.5cm}

%% file: section3.tex
\section{Hierarchical-Temporal Activation Network}
In this section, we propose a flexible and efficient model called Hierarchical-Temporal Activation Network (HTAN), whose task-shared architecture is adaptive with respect to time-variant task relation.

\subsection{Model Architecture}
The architecture of HTAN is composed of a stack of several task adaptive blocks, which are depicted as Figure \ref{Fig:HTAN} (b). We define $x_{\ast}$ and $h_{\ast}$ as the input and output of the block. In each block, a LSTM or attention layer \cite{Vaswani2017NIPS} is introduced as a hidden layer to map the input sequence into a hidden feature. As weights and biases of the hidden layers are the major parameters of deep learning models, HTAN shares the LSTM or attention layer through all the tasks to reduce the model complexity. At time slot $n$, the output feature of the task adaptive block for the input $x_n$ is given as:
\begin{align}
    h_{n} = \mathcal{F}_n^{t}(\widetilde{h}_n=\text{LSTM/Attention}(x_{n};\,\,x_{1:n-1})),
\end{align}
where $x_{1:n-1}$ denotes the input sequence from time slots $1$ to $n-1$. Unlike general deep neural networks that use fixed activation functions, each block of HTAN learns a set of task-specific activation functions \rev{that split the network for different tasks}. For task $t$ at time slot $n$, the activation function is parameterized as Adaptive Piecewise Linear (APL) function \cite{Agostinelli2015ICLR}:
\begin{align}
    \mathcal{F}_n^{t}(\widetilde{h}_n) = \text{ReLU}(\widetilde{h}_n) + \sum_{m=1}^{M}\alpha_{n,m}^t\text{ReLU}(-\widetilde{h}_n + \beta_{n, m}),
\end{align}
where $M$ is the number of the basis functions, $\beta_n=[\beta_{n, 1},\cdots, \beta_{n, M}]\in \mathbb{R}^{M}$ is the bias vector of the basis functions and $\alpha_n^t\in [\alpha_{n, 1},\cdots, \alpha_{n, M}] \in \mathbb{R}^{M}$ is the coordinate vector of the activation function for task $t$. The task adaptive block incorporates two autoregressive LSTMs to compute the time-variant bias vector $\beta_n$ and the task-specific coordinate $\alpha_n^t$ as follows:  
\begin{align}
    \beta_n=\text{LSTM}_{\beta}(\beta_{1:n-1}),\quad \alpha^t_n=\text{LSTM}_{\alpha}(\alpha_n\oplus \beta_n;\,\,\alpha_{1:n-1}),
\end{align}
where $\oplus$ denotes concatenation. The $\text{LSTM}_{\alpha}$ uses $\beta_n$ as input to compute $\alpha^t_n$, as the value of $\alpha_n$ is dependent on the set of basis functions.

\subsection{Discussion}
We design the structure of the adaptive task block based on the following consideration. The activation functions play an important role in introducing non-linearity into the deep neural networks. Given two sequential deep learning models with same weight and bias parameters, their behaviors will be completely different if they use two types of activation functions. Therefore, the similarity between the activation functions is a great component to split the task-specific and shared modules of the sequential DMTL model. If two tasks can share the lower layers of a multi-layer deep neural network, the task-specific activation functions of them will be more similar and vice versa. HTAN can also learn the temporal knowledge splitting by learning time-variant activation functions. Therefore, HTAN can explore both the sharing of hierarchy and the temporal evolution of the hierarchical sharing. 

HTAN also has smaller model complexity than existing soft-sharing methods when there is a large amount of tasks. Compared with the LSTM and transformer block, the adaptive task block in HTAN has two auxiliary LSTM layers. In Section \ref{sec:func}, we also introduce an auxiliary SPDNet for functional regularization. As soft-sharing models define individual modules for each tasks, HTAN has smaller amount of parameters than the soft-sharing methods when the number of trained tasks is larger than $4$.

%% file: section4.tex
\section{Functional Regularization by Modulated SPDnet}\label{sec:func}
The activation functions of HTAN have great flexibility to determine the behavior of the neural network. During the training, HTAN may learn highly different activation functions for two related tasks, which do not benefit DMTL. In order to encourage knowledge sharing for related tasks, it is important to define an efficient regularization by the similarity between the activation functions.

\subsection{Mahalanobis Distance of APL Functions}
In \cite{Liu2019CVPR}, it is shown that the distance between two APL functions can be measure by Mahalanobis distance between the coordinates:
\begin{align}
    	d(\mathcal{F}_{n}^{t_1},\mathcal{F}_{n}^{t_1}) =&\sqrt{ \int(\mathcal{F}_{n}^{t_1}(x)-\mathcal{F}_{n}^{t_2}(x))^2 \mathcal{N}(x)dx} = \sqrt{(\alpha^{t_1}_{n}-\alpha^{t_2}_{n})^T\mathcal{M}(\beta_n)(\alpha^{t_1}_{n} - \alpha^{t_2}_{n})}.\label{eq:dis}
    \end{align}
where $\{t_1, t_2\}$ denote two tasks, $\mathcal{N}(x)$ is a standard normal distribution, and $\mathcal{M}$ is a symmetric positive definite (SPD) matrix induced by $\beta_n$. There is a significant drawback in using Eq. (\ref{eq:dis}) for the time-variant activation functions in HTAN. Eq. (\ref{eq:dis}) is based on the assumption that the input for the two activation functions follow standard normal distribution, which may not be valid in practice. In the hidden blocks of HTAN, the output of the LSTM or attention layer can has different mean and variance in each dimension. The statistics of hidden feature also evolves temporally. Therefore, the assumption of standard normal distribution is generally violated and $\mathcal{M}_0(\beta_n)$ may lead to sub-optimal result. In this section, we propose a deep learning method to learn the time-variant SPD matrices from data.

\subsection{$\beta_n$-SPDNet}
In order to define better similarity metrics for functional regularization, we incorporate a variant of SPDNet \cite{Huang2017AAAI} to compute the SPD matrices for Eq. (\ref{eq:dis}) at each time slot. The variant of SPDNet, which we call it $\beta_n$-SPDNet, is illustrated in Figure \ref{Fig:HTAN} (a). Given an initial SPD matrix $\mathcal{M}_0$, the network consecutively computes transformed SPD matrices $\mathcal{M}_n$ from $\mathcal{M}_{n-1}$ and $\beta_n$ for different time slots.  $\beta_n$-SPDNet consists of a hierarchy with two different hidden layers. The \textbf{BiMap} layer modulated by $\beta_n$ is applied for linear transformation of the input, which is given by  
\begin{align}
    X_k = W_k \Big[X_{k-1} + \text{diag}(\text{ReLU}(V_k \beta_n + b_k))\Big]W_k^T,
\end{align}
where $W_k$ is a weight matrix on Stiefel manifold. The \textbf{ReEig} layer is introduced to insert nonlinearity on the SPDNet: 
\begin{align}
    X_k = U_{k-1}\max\Big(\text{ReLU}(Q_k \beta_n + c_k), \Sigma_{k-1}\Big)U_{k-1}^T,
\end{align}
where $U_{k-1}$ and $\Sigma_{k-2}$ are given by the eigenvalue decomposition of the input.

By training the $\beta_n$-SPDNet with the data, the network can explore the SPD space and generate better $\mathcal{M}_n$ for the Mahalanobis distance, which is important on either analyzing the learned activation functions or regularizing the training process. Therefore, it is necessary to define a proper loss function for $\beta_n$-SPDNet. In the rest of this section, we propose a joint adversarial training method for HTAN and $\beta_n$-SPDNet to achieve  flexible sequential DMTL.

\subsection{Adversarial Training}
In order to embed the temporal-variant task relationship by an interpretable pattern of the activation functions, we propose an adversarial learning method for the core model HTAN and $\beta_n$-SPDNet for regularization. We define the notations $\Phi$ and $\Theta$ as the sets of parameters in HTAN and $\beta_n$-SPDNet respectively. 

The training of HTAN with $L$ task adaptive blocks is achieved by minimizing the following loss function:
\begin{align}
    \Phi^{\ast}=\argmin_{\Phi} \mathcal{L}_{\Phi}, \quad \mathcal{L}_{\Phi} = \sum_{t=1}^{T}\ell_{t}(\Phi) + \lambda  \sum_{l=1}^{L}\sum_{n=1}^{N}||\mathcal{D}(\alpha_{l, n}; \mathcal{M}_{l, n})||_{1, 1}, \label{Eq: loss_phi} 
\end{align}
where $T$ denotes the number of tasks and $\ell_{t}(\Phi)$ is the loss function for the corresponding task. $\lambda$ is the regularization coefficient and $N$ is the length of data sequence. $\mathcal{D}(\alpha_{l, n}; \mathcal{M}_{l, n})\in \mathbb{R}^{T\times T}$ is the square of Mahalanobis distance matrix, where $\mathcal{D}_{i, j}(\alpha_{l, n}; \mathcal{M}_{l, n}) = d^2(\mathcal{F}_{n}^{i},\mathcal{F}_{n}^{j})$. $||\mathcal{D}||_{1, 1}=\sum_{i,j}|\mathcal{D}_{i, j}|$ is the entrywise $L_1$ norm. $\alpha_{l, n}$ and $\mathcal{M}_{l, n}$ are the task-specific coordinate vector and SPD matrix in Eq. (\ref{eq:dis}) respectively.

The training of SPDNet is achieved by minimizing the following loss function:
\begin{align}
    \Theta^{\ast}=\argmax_{\Theta} \mathcal{L}_{\Theta}, \quad \mathcal{L}_{\Theta} = \sum_{i=1}^{T}\log \frac{\exp(d^2(\mathcal{F}_{n}^{i},\mathcal{F}_{n}^{k}))}{\sum_{j\neq k}\exp(d^2(\mathcal{F}_{n}^{i},\mathcal{F}_{n}^{j}))}, \label{Eq: loss_theta} 
\end{align}
where $k$ is the task index whose $\mathcal{F}_{n}^{k}$ is the most different from $\mathcal{F}_{n}^{i}$ (i.e. $d^2(\mathcal{F}_{n}^{i},\mathcal{F}_{n}^{k}) > d^2(\mathcal{F}_{n}^{i},\mathcal{F}_{n}^{j})$, $\forall j\neq k$).

The idea of our proposed adversarial training is introduced as follows.
\rev{In Eq. (\ref{Eq: loss_phi}), the square of distance matrix is incorporated as a regularization term that encourage HTAN to learn similar activation functions for sharing more knowledge across tasks. Optimizing $\mathcal{L}_{\Phi}$ will decrease the distance between task-specific activation functions. When the Mahalanobis distance of some activation functions is already small enough, the model is better paying more attention on the activation function pairs with large difference. Therefore, we train $\beta_n$-SPDNet by $\mathcal{L}_{\Theta}$ to expand the values of the largest distances between the functions. The output SPD matrices will behave like an attention mechanism, which push HTAN to focus more on the dissimilar activation function pairs.}

%% file: section5.tex
\section{Experiments}
In this section, we conduct comprehensive experiments on several challenging applications and compare the performance of HTAN-SPD with the state-of-the-art models.

\subsection{General Language Understanding Evaluation (GLUE)}
The GLUE dataset~\cite{wang2019glue} consists of three types of tasks: single-sentence classification, similarity and paraphrase tasks, and inference tasks, as shown in Table~\ref{tab:glue}.

\textbf{Model Configuration:}

\begin{table}[H]
\caption{Task descriptions and statistics of the GLUE dataset. All tasks are single sentence or sentence pair classification, except STS-B, which is a regression task. All classification tasks have two classes, except MNLI, which has three. \cite{wang2019glue}}
\renewcommand{\arraystretch}{0.8}
\centering
\begin{tabular}{m{1.5cm}m{3.0cm}m{1.5cm}m{1.5cm}m{4.0cm}}
\toprule[1.0pt]
\textbf{Corpus} & \textbf{Task} & $\vert$\textbf{Train}$\vert$ & $\vert$\textbf{Test}$\vert$  & \textbf{Metrics}\\
\midrule[1.0pt]
\multicolumn{5}{c}{Single-Sentence Tasks} \\
\midrule[1.0pt]
CoLA & acceptability  & 8.5k & 1k & Matthews corr. \\
SST-2 & sentiment & 67k & 1.8k & acc. \\
\midrule[1.0pt]
\multicolumn{5}{c}{Similarity and Paraphrase Tasks} \\
\midrule[1.0pt]
MRPC & paraphrase & 3.7k & 1.7k  & acc./F1 \\
STS-B & sentence similarity & 7k & 1.4k & Pearson/Spearman corr. \\
QQP & paraphrase & 364k & 391k & acc./F1 \\
\midrule[1.0pt]
\multicolumn{5}{c}{Inference Tasks} \\
\midrule[1.0pt] 
MNLI & NLI & 393k & 20k & matched/mismatched acc. \\
QNLI & QA/NLI & 105k & 5.4k & acc. \\
RTE & NLI & 2.5k & 3k & acc. \\
WNLI & coreference/NLI & 634 & 146 & acc. \\
\bottomrule[1.0pt]
\end{tabular}
\label{tab:glue}
\end{table}

\begin{table}[H]
	\centering
	\caption{Model Performance on GLUE dataset}\label{sim:table:glue}
	\renewcommand\arraystretch{0.8}
	\small
	\begin{threeparttable}
	\begin{tabular}
		{m{2.0cm}C{1.0cm}C{1.0cm}C{1.0cm}C{1.0cm}C{1.0cm}C{1.0cm}C{1.0cm}C{1.0cm}}\toprule[1.0pt]
		\multirow{2}{*}{\textbf{Model}} &\multicolumn{8}{c}{\textbf{Test Tasks}}  \\
		\cmidrule[1.0pt](lr){2-9}
		&CoLA	&MRPC   &STS-B	&RTE &SST-2	&\rev{QNLI}   &MNLI	&QQP\\
		\midrule[1.0pt]
		BERT \cite{Devlin2019NAACL} &52.1 &88.9/84.8 &87.1/85.8 &66.4  & 93.5 & 90.5 & 84.6/83.4 & ---/71.2\\
		MT-DNN\tnote{$\ddagger$} & 48.0 & 89.6/86.0 & 88.0/87.2 & 75.2 & 93.2 & 91.4 & 83.2/82.3 & 89.2/71.1    \\ 
		MT-DNN \cite{Liu2019ACL} & -- -- & -- -- & -- -- & -- --
		& -- -- & -- -- & -- -- & -- --\\ 
		MT-DNN \cite{Dou2019EMNLP}  &51.7 &89.9/86.3 &87.6/86.8 &75.4
		& -- -- & -- -- & -- -- & -- --\\
		MAML \cite{Dou2019EMNLP}   &53.4 &89.5/85.8 &88.0/87.3 &76.4
		& -- -- & -- -- & -- -- & -- --\\
		FOMAML \cite{Dou2019EMNLP}   &51.6 &89.9/86.4 &88.6/88.0 &74.1
        & -- -- & -- -- & -- -- & -- --\\
		Reptile \cite{Dou2019EMNLP}  &53.2 &90.2/86.7 &88.7/88.1 &77.0
		& -- -- & -- -- & -- -- & -- --\\
		\midrule[1.0pt]
		\bfseries HTAN & 54.8 & 91.3/87.9 & 89.9/89.0 & 79.3 & 95.8 & 93.7 & 86.5/84.7 & 91.2/87.5 \\
		\bottomrule[1.0pt]
	\end{tabular}
	\begin{tablenotes}
    \item[$\ddagger$] Our training results;
    \end{tablenotes}
	\end{threeparttable}
\end{table}

%% file: section6.tex
\section{Conclusion}
Deep learning has been widely used in many tasks, such as image captioning~\cite{yang-etal-2021-journalistic,Yang2020Fashion,reformer}, image to image translation~\cite{yang2018cross}, neural machine translation~\cite{yang2019latent}, robotics~\cite{xuewen14,feng2015}, fashion recommendation~\cite{xuewen_mm20}, OCR~\cite{yang_license} etc. In this paper, we investigated a brand-new framework for multi-task learning on the spatial-temporal data. The proposed framework can learn a learnable activation function which is flexible and efficient. Through extensive experiments, we show that our method outperformed the other state-of-the-art methods. In the future, we could do more experiments especially on the high-dimensional data to prove its performance. The tasks include neural machine translation, image super-resolution, and speech recognition.

%% file: section_appendix.tex
\appendix
\section{GLUE Tasks}
GLUE contains nine language understanding tasks, which cover a broad range of domains, data quantities, and difficulties. 

\textbf{Single-Sentence Classification.} 
This type of tasks ask the model to make a prediction given a single sentence. 
For \textbf{CoLA}, the model has to predict whether a given sentence is a grammatical English sentence. 
Matthews correlation coefficient~\cite{mat1975} is used as the evaluation metric, which evaluates performance on unbalanced binary classification and ranges from $-1$ to $1$, with $0$ being the performance of uninformed guessing.
The goal of \textbf{SST-2} is to determine the sentiment of a sentence as positive or negative.

\textbf{Similarity and Paraphrase Tasks} ask the model to determine whether or to what extent two given sentences are semantically equivalent to each other. 
The \textbf{MRPC} and the \textbf{QQP} are two classification tasks, while the \textbf{STS-B} is a repression task and requires the model to output a similarity score from $1$ to $5$.
Pearson and Spearman correlation coefficients are used as the evaluation metrics for the \textbf{STS-B} task.

\textbf{Inference Tasks}.
Given a premise sentence and a hypothesis sentence, the \textbf{MNLI} task is to predict whether the premise entails the hypothesis (\textit{entailment}), contradicts the hypothesis (\textit{contradiction}), or nither (\textit{neutral}).
Similar to \textbf{MNLI}, \textbf{RTE} uses a two-class split, where \textit{neutral} and \textit{contradiction} are collapsed into \textit{not\_entailment}.
\textbf{QNLI} is converted from a question-answering task into a sentence pair classification task, where the objective is to determine whether the context sentence contains the answer to the question.
\textbf{WNLI} is to predict if the sentence with the pronoun substituted is entailed by the original sentence. 
Because the test set is imbalanced and the development set is adversarial, so far none of the proposed models could surpass the performance of the simple majority voting strategy. 
Therefore, we do not use the WNLI dataset in this paper.
Other papers adopted the same strategy~\cite{Liu2019ACL,Dou2019EMNLP}.